\pdfoutput=1
\documentclass[reqno,11pt]{article}
\usepackage{amsthm}
\usepackage{amssymb,amsmath}
\usepackage{xypic}
\usepackage[colorlinks = true,
            linkcolor = blue,
            urlcolor  = blue,
            citecolor = blue,
            anchorcolor = blue]{hyperref}
\usepackage{bbm}
\usepackage{pbox}

\usepackage{tikz}
\usepackage[font=scriptsize]{caption}
\usepackage{algorithm}
\usepackage{algorithmic}

\usepackage[affil-it]{authblk}

\title{Autoencoding topology} 
\author{Eric O. Korman\thanks{eric.korman@gmail.com}}
\date{\vspace{-4ex}}

\renewcommand{\phi}{\varphi}
\newcommand{\eps}{\varepsilon}

\newcommand{\U}{\mathcal U}
\newcommand{\X}{\mathcal{X}}
\newcommand{\Z}{\mathcal{Z}}
\newcommand{\D}{\mathcal{D}}

\newcommand{\R}{\mathbb{R}}
\newcommand{\E}{\mathbb E}

\newtheorem{remark}{Remark}

\usepackage{tikz}
\usetikzlibrary{shapes.geometric, arrows, automata, fit, positioning}

\tikzstyle{tnode} = [rectangle, rounded corners, minimum width=.75cm, minimum height=.75cm,text centered, draw=black]
\tikzstyle{node} = [circle,  minimum width=.2cm, minimum height=.2cm, draw=black]
\tikzstyle{arrow} = [thick,->,>=stealth]
\tikzstyle{edge} = [thick]
\tikzstyle{dashedarrow} = [dashed,->,>=stealth]

\definecolor{538g}{RGB}{109,144,79}
\definecolor{538b}{RGB}{52,138,189}
\definecolor{538r}{RGB}{226,74,51}

\begin{document}
\maketitle
\abstract{The problem of learning a manifold structure on a dataset is framed in terms of a generative model, to which we use ideas behind autoencoders (namely adversarial/Wasserstein autoencoders) to fit deep neural networks. From a machine learning perspective, the resulting structure, an atlas of a manifold, may be viewed as a combination of dimensionality reduction and ``fuzzy" clustering.}

\section{Introduction}
A $d$-dimensional submanifold $\X$ of $\R^n$ is specified by an \textit{atlas}, which is a collection of maps (called \textit{charts}) that give local, smooth identifications of $\X$ with open subsets of $\R^d$.  In this work we take ``manifold learning" literally and give a technique for fitting an atlas to an (i.i.d.) sample of points from $\X$. This is achieved by viewing an atlas as a generative model, to which we fit neural networks using the technique of adversarial autoencoders (AAE) \cite{AAE}, which is a special case of the more general framework of Wasserstein autoencoders with a GAN-based penalty (WAE-GAN) \cite{WAE}.

\begin{figure}[h]
\centering
\captionsetup{width=.7\linewidth}
\begin{tikzpicture}[scale=.65]

\draw[thick] (0,0) circle (1.5cm);
\draw[538b, thick] (1.65,0) arc (0:130:1.65cm);
\draw[538r, thick] (-.9,1.55884572681) arc (120:250:1.8cm);
\draw[538g, thick] (-.975,-1.68874953738) arc (240:370:1.95cm);

\end{tikzpicture}   \hspace{30pt}  \raisebox{0.2\height}{\begin{tikzpicture}[node distance=1.5cm]
\node (0) [node, color=538b, fill=538b] {};
\node (1) [node, color=538r, fill=538r, below of=0, left of=0] {};
\node (2) [node, color=538g, fill=538g, below of=0, right of=0] {};

\draw [edge]  (0) -- (1);
\draw [edge]  (0) -- (2);
\draw [edge]  (1) -- (2);

\end{tikzpicture}}
\caption{An atlas of the circle $S^1 \subset \R^2$ and its \v Cech nerve. Each node of the graph represents a chart and there is an edge between two nodes if the corresponding charts have a non-empty intersection. Note that the graph itself is homotopy equivalent to $S^1$.}
\label{S1-atlas}
\end{figure}
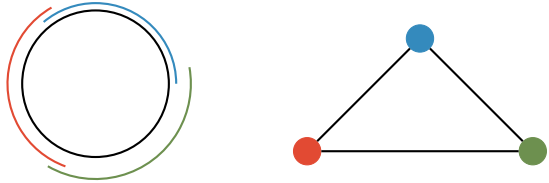

This work has both theoretical and practical motivations.  From a mathematical perspective, an interesting question is how the topology or homotopy type of a space can be recovered from a sample of points. Besides an atlas containing all of the topological information of a manifold, in the special case that it forms a \textit{good cover} (i.e. the intersection of any collection of charts is contractible) the homotopy type of the manifold can be recovered from a simple combinatorial object, the \v Cech nerve, that keeps track of the intersections between the various charts. An example for the circle is given in figure \ref{S1-atlas}. For embedded submanifolds of $\R^n$, one way of encouraging an atlas to be a good cover is by taking a large number of charts, each of which is the restriction of a linear map $\R^n \to \R^d$. This corresponds to the encoder networks being single, linearly activated layers.

On the other hand, from the perspective of unsupervised representation learning, the structure of an atlas gives a simultaneous generalization of dimensionality reduction (the case of an atlas that consists of a single chart) and clustering (in the case of an atlas with charts that do not overlap). Our autoencoder based approach is particularly robust since we have a probabilistic encoder, so that we can map new data to the smaller dimensional latent space, as well as a probabilistic decoder, so that we can generate synthetic data from the latent space.

\subsection{Related work and background}
For another approach to fitting an atlas to a sample of points, see \cite{ATLAS}. An atlas determines in particular a simplicial complex via the \v Cech nerve construction. Similar complexes form the basis of ``Topological Data Analysis", whose methods include persistent homology and the MAPPER algorithm. See \cite{TDA} and the references therein. 

Good references for the topological objects we discuss (e.g. manifolds, simplicial complexes, nerves, and good coverings) include \cite{BottTu} and \cite{Hatcher}.

\section{Setup}
Let $\X \subset \R^n$ be an embedded $d$-dimensional submanifold of $\R^n$ and fix a positive integer $k$. From a sample of i.i.d. points $\X_{sample} = \{x_1,\ldots,x_N\} \subset \X$ we wish to infer an atlas of $\X$ consisting of $k$ coordinate charts diffeomorphic to the open set $\Z := (-1,1)^d \subset \R^d$. Specifically, we seek to find maps $\phi_j : \Z \to \X$ for $j=1,\ldots,k$ such that
\begin{enumerate}
\item Each $\phi_j$ is a diffeomorphism onto its image.
\item For every $x \in \X$ there exists some $j$ such that $x \in \phi_j(Z)$.
\end{enumerate}
\begin{remark}\label{linearity-remark}
By taking $k$ to be sufficiently big, we can always find an atlas such that $\phi_j^{-1} : \R^n \supset \phi_j(\Z) \to \Z$ is the restriction of a linear map $\R^n\to \R^d$ (note that this does not mean that $\phi_j$ itself is linear). A mathematical benefit of forcing the $\phi_j^{-1}$'s to be linear is that it encourages the atlas to form a good cover.
\end{remark}

\begin{figure}
\centering
\begin{tikzpicture}

\node (latent1) [tnode] {$\Z$};

\node (latent2) [tnode, below=.5cm of latent1] {$\Z$};

\node (latentk) [tnode, below=1cm of latent2] {$\Z$};

\node (X) [tnode, right=1cm of latent2] {$\X$};

\draw [arrow] (latent1) -- (X);

\draw [arrow] (latent2) -- (X);

\draw [arrow] (latentk) -- (X);

\draw[dashedarrow] (X) to [out=100,in=0] (latent1);

\draw[dashedarrow] (X) to [out=150,in=30] (latent2);

\draw[dashedarrow] (X) to [out=270,in=0] (latentk);

\path (latent2) -- node[auto=false]{\vdots} (latentk);
\end{tikzpicture}
\captionsetup{width=.75\linewidth}
\caption{Schematic of the generative model. The solid lines denote $p(X \mid J, Z)$ while the dotted lines denote the posteriors $p(Z \mid J, X)$.}
\label{schematic}
\end{figure}
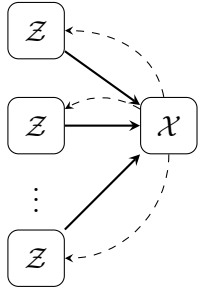

We view this problem in the framework of generative models, where the latent space is 
\[
\Z \times\underbrace{\{1, \ldots, k\}}_{:= \mathcal J}
\]
with the uniform prior $p(Z, J)$ \footnote{We will use the convention of upper-case calligraphic fonts (e.g. $\X$) for a space, upper-case letters for a random variable valued in that space (e.g. $X$) and lower-case letters for a point in the space (e.g. $x$).} and
\[
p(X \mid J=j, Z) = \mathbbm 1_{X = \phi_j(Z)}.
\]
The posterior conditioned on $J$, $p(Z \mid X=x, J=j)$, is deterministic via $\phi_j^{-1}(x)$.  This is summarized schematically in figure \ref{schematic}.

\section{Fitting neural networks}\label{sec-fitting-neural-networks}
We use $q$ to represent an approximation to the true probability distribution, $p$. The autoencoder model will be fit using three types of neural networks:
\begin{enumerate}
\item $k$-many \textit{encoder} networks $q(Z \mid J=1, X), \ldots, q(Z\mid J=k, X)$.
\item $k$-many \textit{decoder} networks $q(X \mid J=1, Z), \ldots, q(X\mid J=k, Z)$.
\item A \textit{chart membership} network $q(J \mid X)$.
\end{enumerate}
We will use deterministic encoder and decoder networks, which is consistent with our interpretation in terms of charts. However, much of what we do is applicable to non-determinstic encoders/decoders, in case we wish to model a noisy manifold. We will abuse notation slightly and reuse $\phi_j$ from the last section to denote the approximate $j$th chart corresponding to $q$: 
\[
q(X\mid J=j, Z) = \mathbbm 1_{X = \phi_j(Z)}
\]
The function $\psi_j : \X \to \Z$ satisfying
\[
q(Z\mid J=j, X) = \mathbbm 1_{Z = \psi_j(X)}
\]
is to approximate the inverse of $\phi_j$ (when restricted to the image of $\phi_j$).

Following the techniques of \cite{WAE} and \cite{AAE}, we seek to minimize the 2-Wasserstein distance between the distribution $p(X)$ (which we estimate using the training data) and the distribution coming from pushing the uniform distribution on the latent space $\Z \times \mathcal J$ to $\X$ via the decoder. This amounts to trying to simultaneously minimize
\begin{equation}
\E_{p(X)} \E_{q(Z, J\mid X)} \mathcal C(X,\phi_J(Z)) \approx \frac{1}{N} \sum_{i=1}^N \sum_{j=1}^k q(j \mid x_i) \mathcal ||x_i - \phi_j(\psi_j(x_i))||^2 \label{AE-objective}
\end{equation}
and
\[
\mathcal D(q(Z, J), p(Z, J)),
\]
where $\mathcal \D$ is a divergence, which we will take to be the Jensen-Shannon divergence. 

The distribution $q(Z,J \mid X)$, which is the product of the deterministic part $q(Z \mid X,J)$ and the probabilistic part $q(J \mid X)$, gives the marginal distribution
\begin{equation} \label{GAN-generator}
q(Z,J) = \int q(Z,J \mid X=x) p(x) dx \approx \frac{1}{N} \sum_{i=1}^N q(J \mid X=x_i) q(Z \mid J, X = x_i),
\end{equation}
which is matched to the prior $p(Z, J)$ by using adversarial training to minimize the Jensen-Shannon divergence between the two distributions.  We thus introduce a discriminator 
\[
D  = (D_1,\ldots, D_k) : \Z \times \mathcal J \to \R
\]
whose goal is to classify points as coming from the prior distribution.  Note that the generator distribution of the GAN is given by (\ref{GAN-generator}).

$D$ is trained to maximize the function
\begin{align}
\E_{p(Z,J)} &\log D(Z,J) + \E_{q(Z,J)} \log(1 - D(Z,J)) \nonumber\\
&= \E_{p(Z,J)} \log D(Z,J) + \E_{p(X)} \sum_{j=1}^k q(j \mid X) \log(1 - D_j(\psi_j(X))) \nonumber\\
&\approx \E_{p(Z,J)} \log D(Z,J) + \frac{1}{N} \sum_{i=1}^N \sum_{j=1}^k q(j \mid x_i) \log(1 - D_j(\psi_j(x_i)))   \label{D-objective}
\end{align}
while $G$ is trained to minimize it. As is common in GAN training \cite{GAN, WAE}, we instead train $G$ to maximize
\begin{equation}
\frac{1}{N} \sum_{i=1}^N \sum_{j=1}^k q(j \mid x_i) \E_{q(Z \mid j,x_i)} \log D_j(\psi_j(x_i)). \label{G-objective}
\end{equation}
The case $k=1$ gives exactly an AAE. For higher values of $k$, this amounts to simultaneously training $k$-many AAEs along with the network $q(J \mid X)$, where now the total reconstruction loss and the ``false positive" part of the discriminator loss is the sum of the losses of each AAE, weighted by $q(J\mid X)$. 

The algorithm thus begins by initializing encoder networks $\psi_1, \ldots, \psi_k$, decoder networks $\phi_1,\ldots, \phi_k$, a chart membership network $q(J\mid X)$, and discriminator networks $D_1, \ldots, D_k$. We sample a mini-batch $\X_{batch}$ from $\X_{data}$ and first update the encoder networks, decoder networks, and chart membership networks using gradient descent to minimize the reconstruction error (\ref{AE-objective}). We then sample from the uniform prior $p(Z,J)$ and use gradient descent to update the discriminator networks to maximize (\ref{D-objective}) on $\X_{batch}$. Finally, we update the encoder networks again to maximize (\ref{G-objective}) on $\X_{batch}$. 

\subsection{Using linear encoders}
By remark \ref{linearity-remark}, it is reasonable to take each encoder network $\psi_j$ to be a single, linear layer as long as $k$ is chosen to be sufficiently large. Using such a simple network for the encoders adds interpretability to the model  and also means we do not have to worry about what the structure of the network should be. Further, recent work has highlighted possible issues that may arise from using deterministic and non-linear encoders, especially when the latent space and intrinsic dimension are not equal \cite{WAE-latent}.
\section{Topological inferences} \label{sec-topology}

\subsection{Estimating dimension}
\begin{figure}[h]
\centering
\captionsetup{width=.75\linewidth}
\includegraphics[scale=.21]{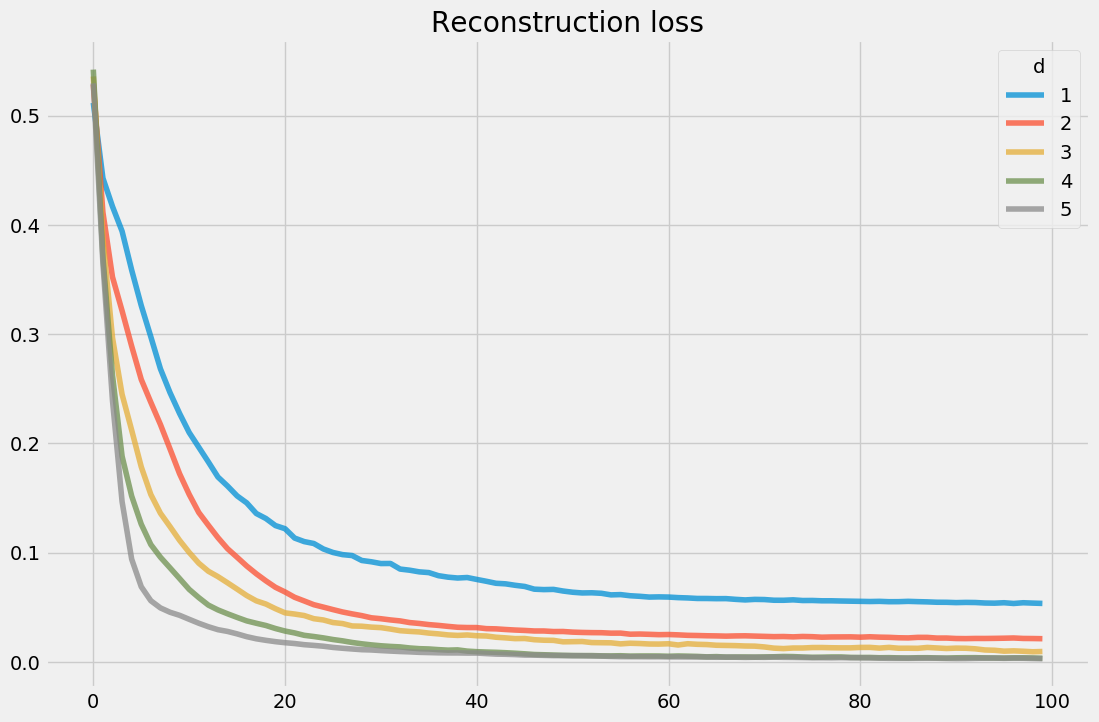} \includegraphics[scale=.21]{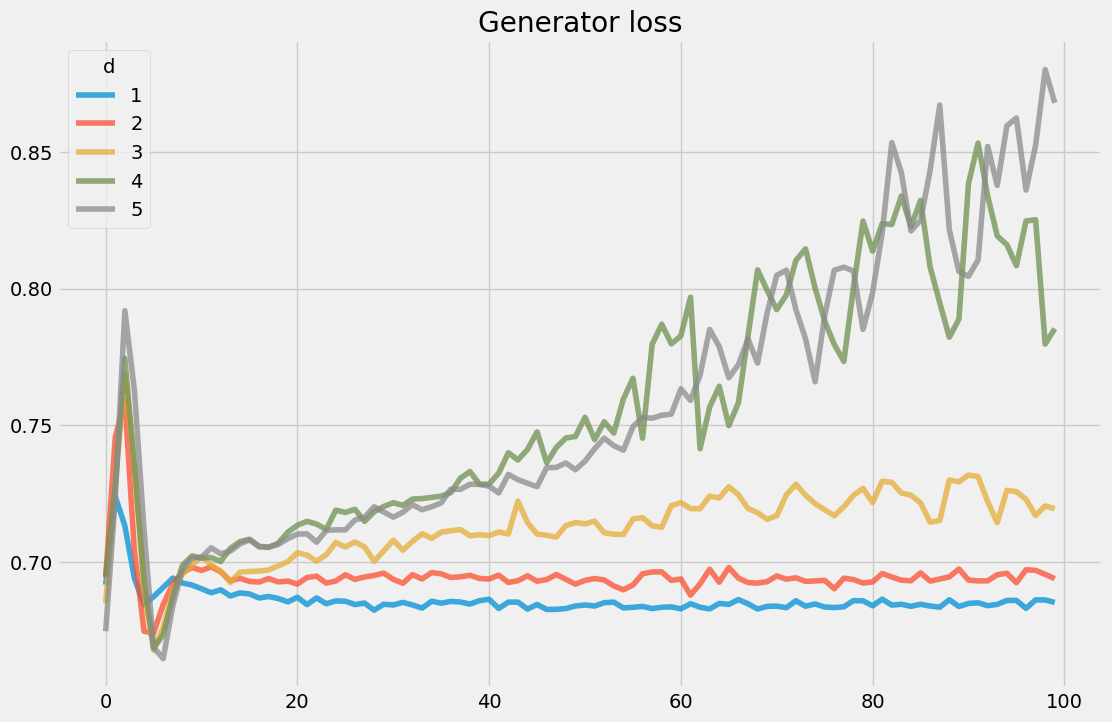}
\caption{The reconstruction and generator losses, over training epochs, for fitting atlases to $\mathbb T^3$ for various choices of $d$.}
\label{T3-losses}
\end{figure}

One may hope to extract the intrinsic dimension of $\X$ by studying the losses as the parameter $d$ varies. While increasing $d$ will decrease the reconstruction error in general, for values of $d$ larger than the actual dimension of $\X$ it is expected that the discriminator and generator losses will drop below and above, respectively, their ideal values of $\log 4$ and $\log 2$. This is because the local generators (i.e. encoders) will be unable to make something of lower dimension appear higher (as opposed to generating lower dimensional points from higher dimensional ones). This will be especially true when we take the generators to be linear. As an example we fit atlases (with linear encoders) of varying dimension to the 3-torus $\mathbb T^3 = S^1\times S^1\times S^1$ embedded into $\R^6$ via three copies of the usual embedding $S^1\hookrightarrow \R^2$. Plots of the losses are in figure \ref{T3-losses}. We caution that general instability of GAN training means that one must be careful when making inferences based on losses. 

\subsection{The \v Cech nerve}
An atlas of $\X$ gives, in particular, an open cover of the manifold, i.e. a collection of open subsets $\mathcal U = \{U_\alpha\}$ of $\X$ whose union is all of $\X$. The \textit{\v Cech nerve} of $\U$ is the simplicial complex whose $\ell$-simplicies are
\[
\{[\alpha_0, \ldots, \alpha_\ell] \mid U_{\alpha_0} \cap \cdots\cap U_{\alpha_\ell} \ne \emptyset \}.
\]
Under favorable circumstances, e.g. if the cover is a \textit{good cover} (which means all of the intersections of elements of $\U$ are contractible), the \v Cech nerve is homotopy equivalent to $\X$. In our probabilistic setup, we use $q(J \mid X)$ to measure how much two charts overlap. We discuss two methods, each of which uses a hyperparameter, $\eps$. \\

\textit{Method 1:} The simplest way is to declare that charts $U_{i_0}, \ldots, U_{i_k}$ overlap if there exists some $x \in \X_{sample}$ such that $q(j_0 \mid X=x), \ldots, q(j_k \mid X=x)$ are all greater than some tolerance $\eps$. \\

\textit{Method 2:} More robustly, the quantity $\E_{X\sim p(X\mid J=j_0)} p(J=j_1 \mid X)$ is a measure of how much chart $j_0$ is contained in $j_1$. We may then take
\[
u_{j_0j_1} = \frac{1}{2} \left(\E_{X \sim p(X \mid J=j_0)} p(J = j_1 \mid X) + \E_{X \sim p(X \mid J=j_1)} p(J = j_0 \mid X) \right) 
\]
as a measure of how much charts $j_0$ and $j_1$ overlap. Using $q(J\mid X)$ and the sample distribution on $X$, this is approximated as
\begin{equation}
u_{j_0j_1} \approx \frac{1}{2}\left( \frac{1}{\sum_i  q(j_0 \mid x_i)} + \frac{1}{\sum_i q(j_1 \mid x_i)} \right) \sum_{i=1}^N q(j_0 \mid x_i) q(j_1 \mid x_i). \label{doubleoverlap}
\end{equation}
Then we consider $U_{j_0}$ and $U_{j_1}$ to overlap if $u_{j_0 j_1} > \eps$.

In general, we bootstrap this to higher degree intersections as follows. Suppose all of the $\ell$-fold intersections between sets from $\{U_{j_0}, \ldots, U_{j_\ell}\}$ are deemed to be nonempty.  Then we normalize the function $p(x \mid J=j_{r_1}) \cdots p(x \mid J = j_{r_\ell})$, which we denote by $p_{j_{r_1}\cdots j_{r_\ell}}(x)$, to serve as a proxy for the pdf on $U_{j_{r_1}} \cap \cdots \cap U_{j_{r_\ell}}$ and use
\begin{align*}
u_{j_0\cdots j_\ell} &= \frac{1}{\ell+1} \sum_{k=0}^\ell \E_{X \sim p_{j_0\cdots \widehat{j_k} \cdots j_{\ell}}(X)} p(J = j_k \mid X) \\
&\approx \frac{1}{\ell + 1} \sum_{k=0}^\ell \left( \frac{1}{\sum_i q(j_0 \mid x_i) \cdots \widehat{q(j_k \mid x_i)} \cdots q(j_\ell \mid x_i) }\right)  \sum_{i=1}^N q(j_0 \mid x_i) \cdots q(j_\ell \mid x_i)
\end{align*}
as a measure for how much $U_0, \ldots, U_\ell$ overlap, where $\hat{~}$ denotes omission.

Using either method, for any value of $\eps \in (0,1)$ we get a simplicial complex and we can study the homology of these complexes as $\eps$ varies. This is very similar, and motivated by, the construction of barcodes in persistent homology \cite{TDA}. 

\begin{figure}
\centering
\captionsetup{width=.85\linewidth}
\includegraphics[scale=.4]{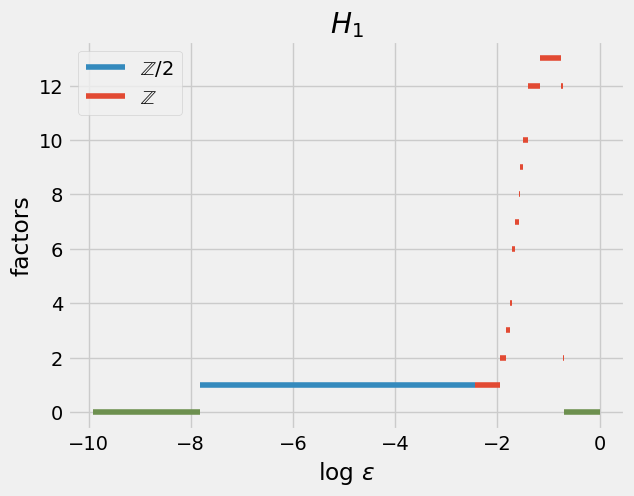} \includegraphics[scale=.4]{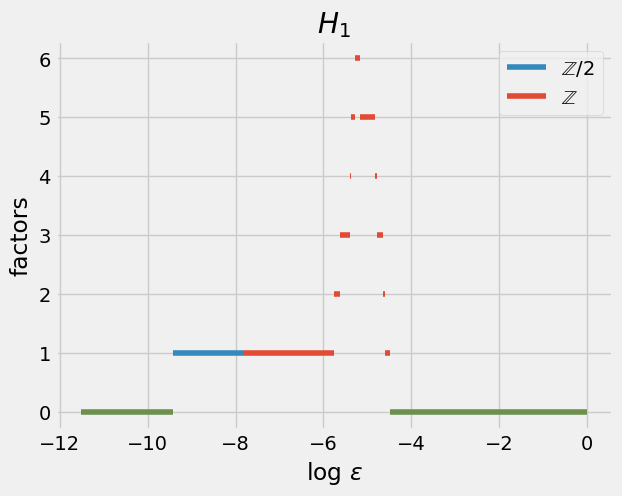}

\includegraphics[scale=1]{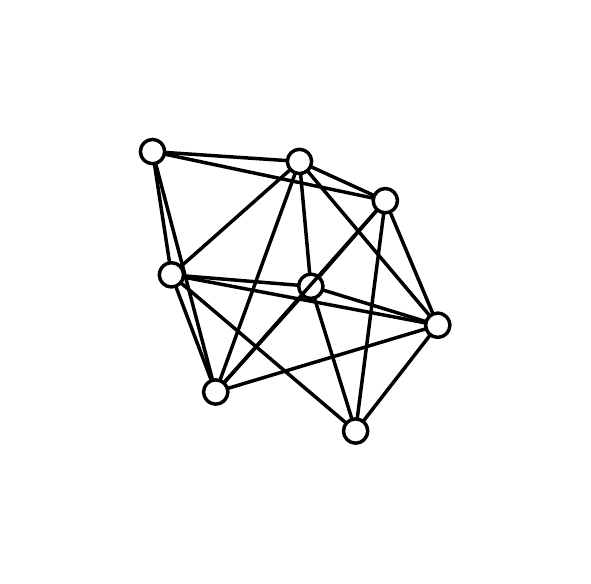}
\caption{Plots of $H_1$ vs. $\log\eps$ for the two methods discussed (left is method 1, right is method 2) and a sample one-skeleton for method 1 with $\log \eps = -5$. For the blue points $H_1 \simeq \mathbb Z/2$ while for the red points $H_1 \simeq \mathbb Z^{factor}$.}
\label{RP2-oneskel}
\end{figure}

As an example, we consider the real-projective plane, $\R P^2 := S^2 / (x \sim -x)$, which has a non-trivial first homology group $H_1(\R P^2; \mathbb Z) = \mathbb Z/2$. We fit an atlas using $d=2$ and $k=8$ by embedding $\R P^2$ into $\R^4$ via the map
\[
\R^3 \supset S^2 \to \R^4, ~~ (x,y,z) \mapsto (x^2 - y^2, xy, xz, yz)
\]
and sampling 10,000 points uniformly.  Figure \ref{RP2-oneskel} shows how the homology varies over $\log \eps$.

\pagebreak

\subsection{MNIST}
We apply our techniques to the MNIST dataset, consisting of 70,000 28x28 pixel images, and consider both cases of non-linear and linear encoders. With non-linear encoders we achieved good results using 15 charts, while for linear encoders we found that 40-120 charts were needed.

\begin{figure}
\centering
\includegraphics[scale=.25]{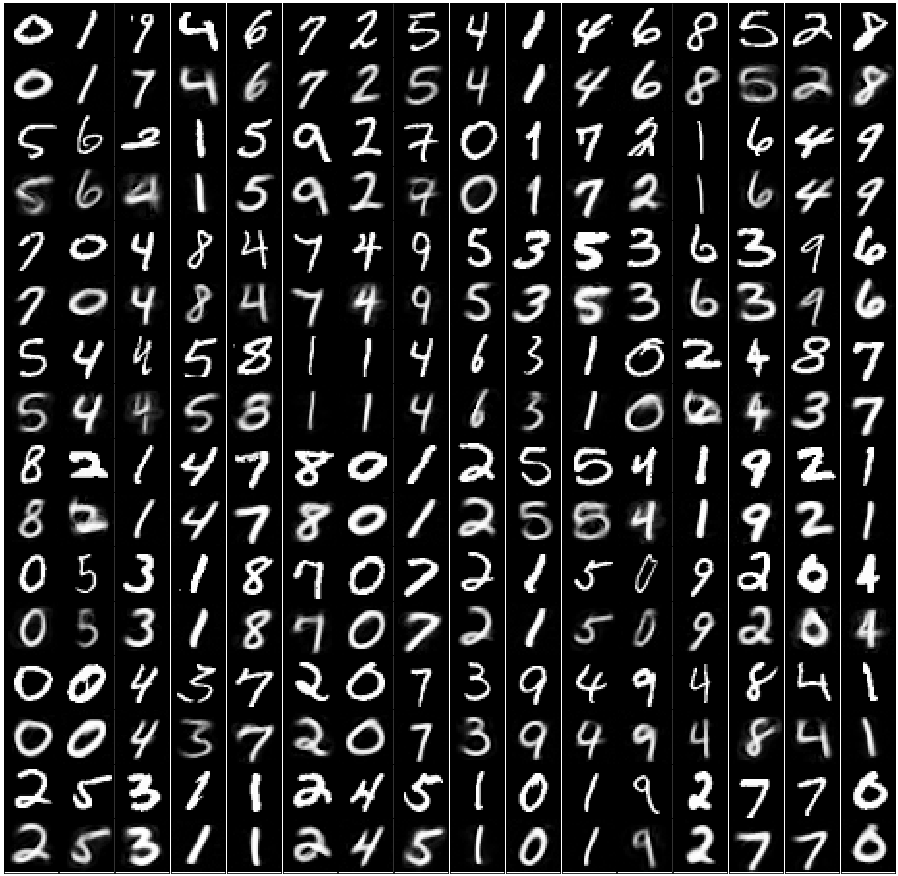} \includegraphics[scale=.25]{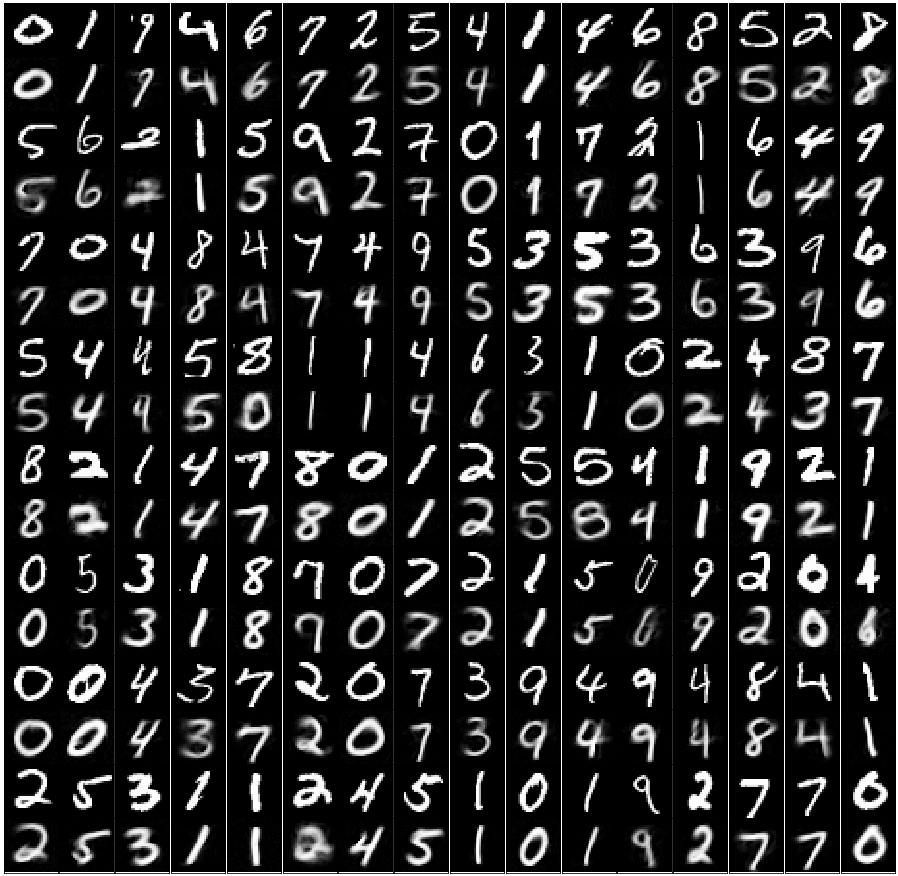}
\caption{Reconstructions. Odd rows are actual datapoints and even rows are their reconstructions. The left image comes from an atlas with $d=11$, 15 charts, and non-linear encoders. The right image comes from an atlas with $d=12$, 40 charts, and linear encoders.}
\label{reconstructions}
\end{figure}

\begin{figure}
\centering
\includegraphics[scale=.25]{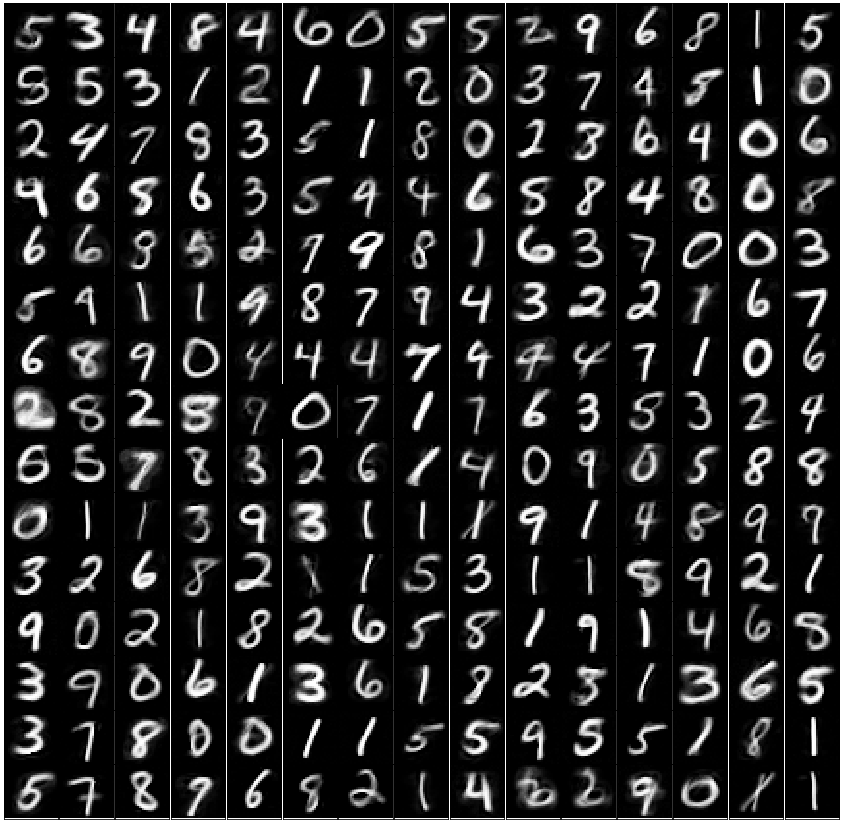} \includegraphics[scale=.25]{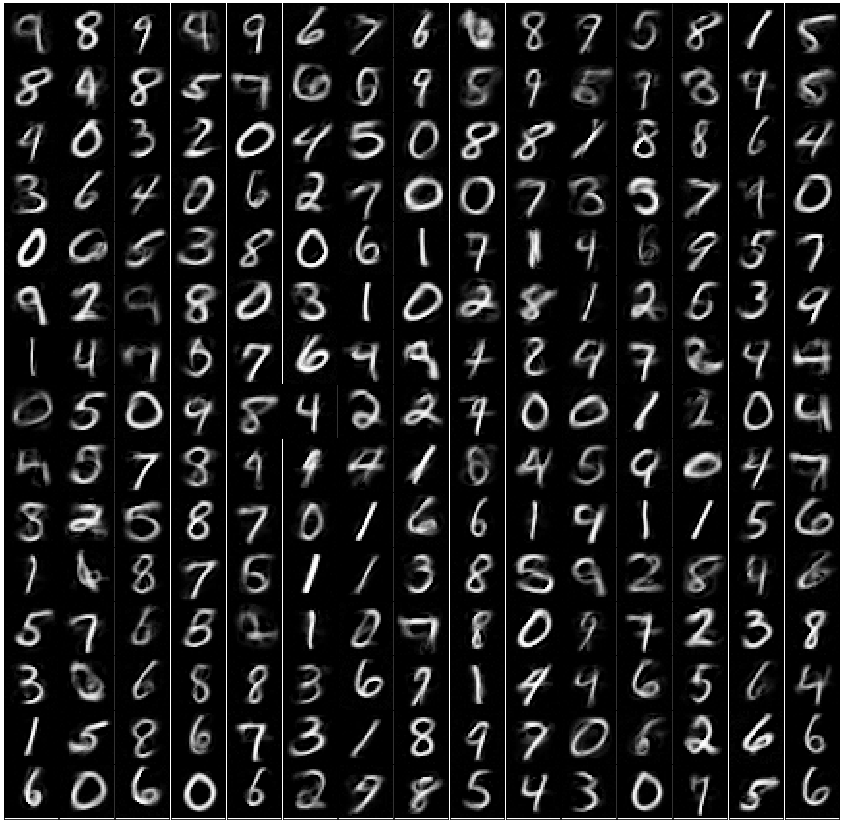}
\caption{Generated images. The left image comes from an atlas with $d=11$, 15 charts, and non-linear encoders. The right image comes from an atlas with $d=12$, 40 charts, and linear encoders.}
\label{generated-images}
\end{figure}

In figures \ref{reconstructions} and \ref{generated-images} are examples of reconstructions and generated images, obtained by sampling $j$ uniformly from $\{1,\ldots, k\}$ and then applying $\phi_j$ to a uniform sample from $(-1,1)^d$. 

\begin{figure}
\centering
\includegraphics[scale=.3]{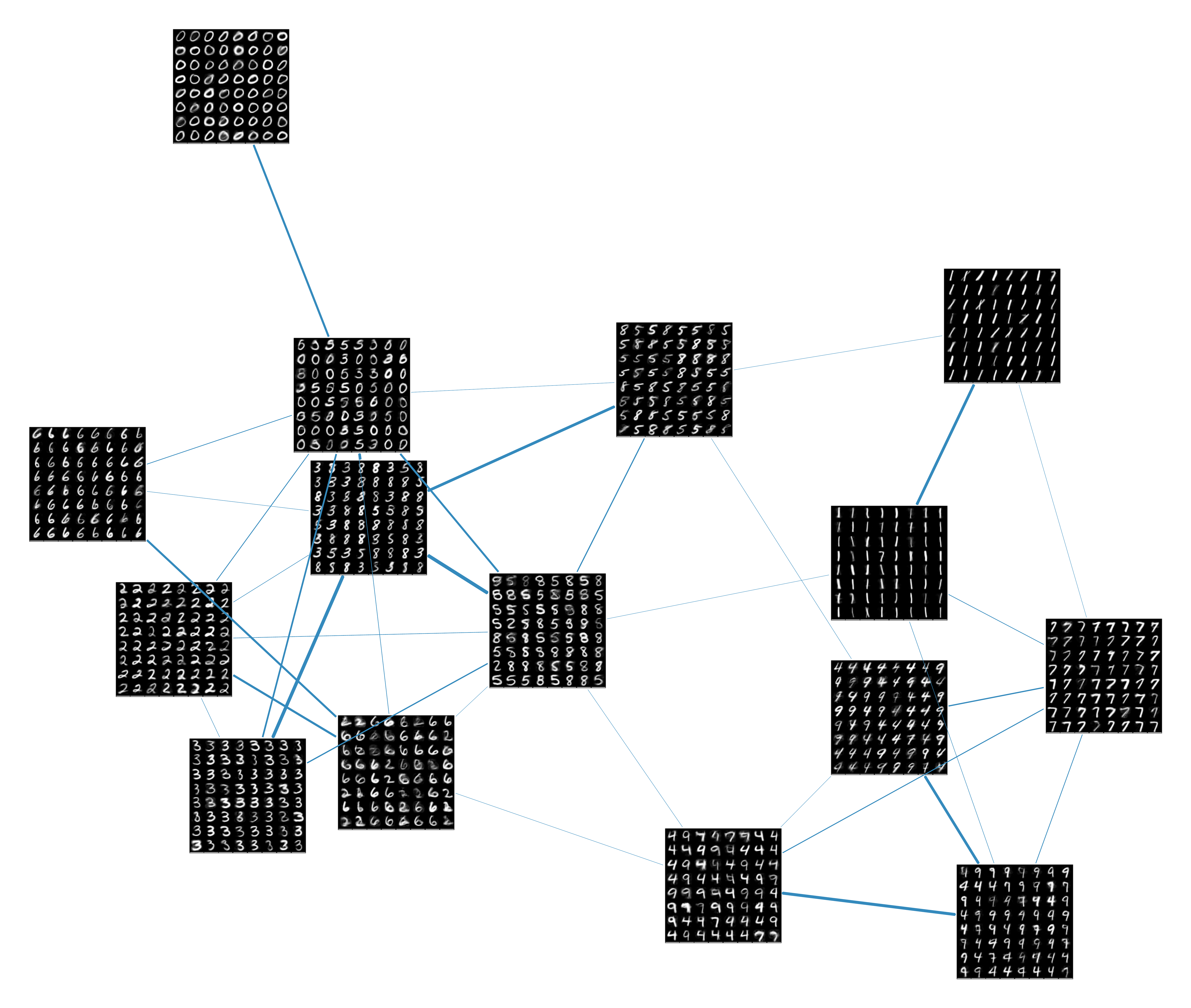}
\caption{Generated images from each of the 15 charts for an atlas with $d=11$.  An edge denotes overlap and thicker lines correspond to more substantial overlap.}
\label{oneskel-d11}
\end{figure}

\begin{figure}
\centering
\includegraphics[scale=.3]{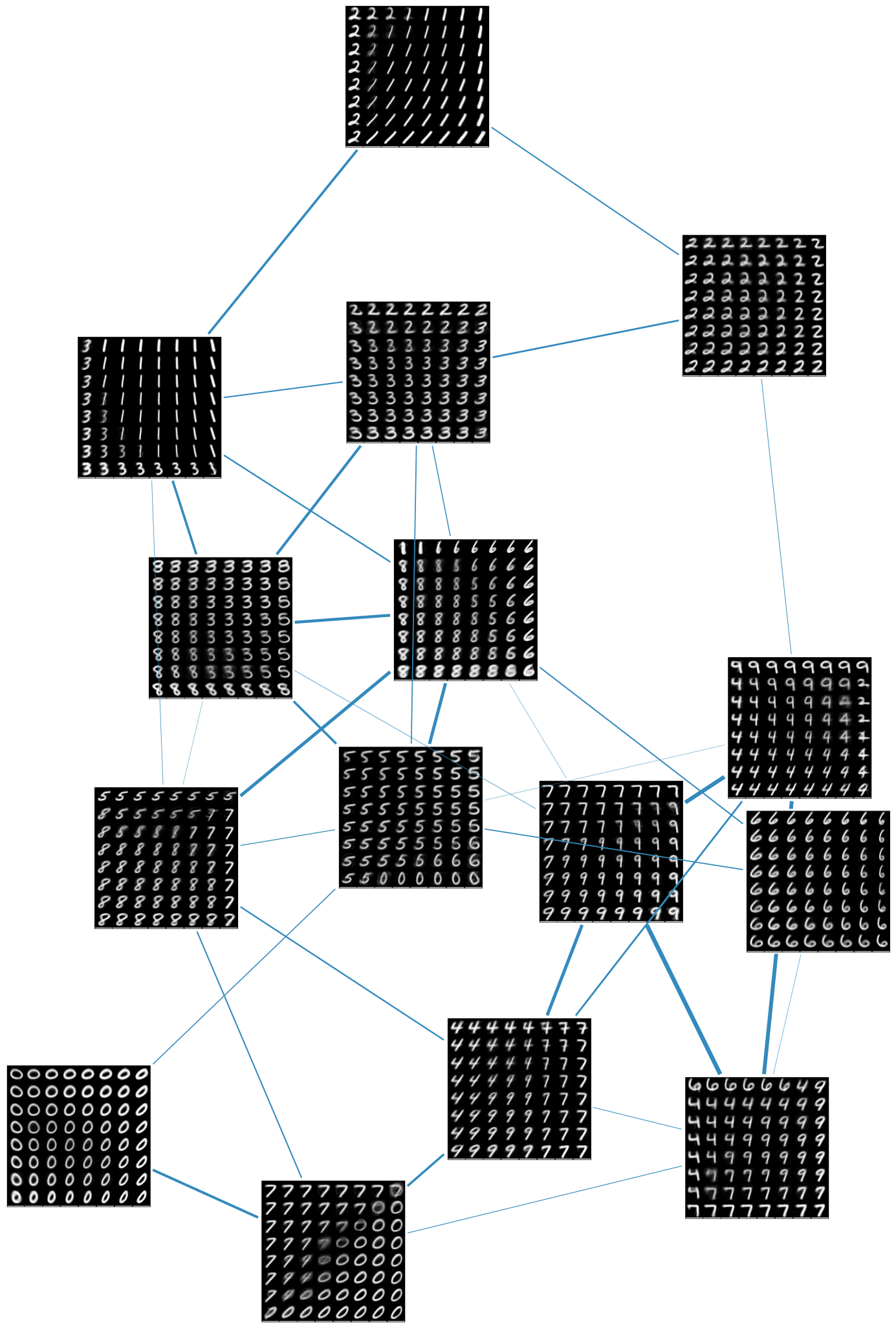}
\caption{Generated images from each of the 15 charts for an atlas with $d=2$ (latent points were chosen evenly across an $8\times8$ grid).  An edge denotes overlap and thicker lines correspond to more substantial overlap.}
\label{oneskel-d2}
\end{figure}

We also look at visualizations of the one-skeletons (figures \ref{oneskel-d11} and \ref{oneskel-d2}). For $d > 2$ we uniformly draw 64 samples from each latent chart $(-1,1)^d$ and apply the decoders to generate images. For $d=2$ we do a similar thing except instead of generating from random samples we apply each chart's decoder to the set $\{0, 1/8, \ldots, 7/8\} \times \{0, 1/8, \ldots, 7/8\}$. We weigh each of the double overlaps using (\ref{doubleoverlap}) and draw the corresponding edges for the top third, with thickness proportional to $u_{j_0j_1}$.

\pagebreak 
\bibliographystyle{plain}
\bibliography{biblio}

\appendix
\section{Experimental details}
For all training we used the RMSprop optimizer with a learning rate of $10^{-3}$ and a mini-batch size of 128. For the geometric examples of $\mathbb T^3$ and $\R P^2$ in section \ref{sec-topology}, all encoder networks were linear and the decoder networks consisted of two 16-dimensional hidden layers with relu activations, followed by a final $\tanh$ activation. The chart membership network, $q(J \mid X)$ consisted of one 16-dimensional hidden layer followed by an output layer with softmax activation.

For the MNIST examples, the atlas with linear encoders had decoder and discriminator networks with two 64-dimensional hidden layers with relu activations. The chart membership network consisted of a single 64-dimensional hidden layer with relu activation. For the atlas with non-linear encoders, the chart membership network and all of the encoders shared two 128-dimensional hidden layers with relu activations. The final layer of each encoder had a $\tanh$ activation. The decoder and discriminator networks had two 128-dimensional hidden layers with relu activations and final layers with $\tanh$ and softmax activations, respectively.

\end{document}